\documentclass[conference]{IEEEtran}
\IEEEoverridecommandlockouts
\usepackage{cite}
\usepackage{amsmath,amssymb,amsfonts}
\usepackage{algorithmic}
\usepackage{graphicx}
\usepackage{textcomp}
\def\BibTeX{{\rm B\kern-.05em{\sc i\kern-.025em b}\kern-.08em
    T\kern-.1667em\lower.7ex\hbox{E}\kern-.125emX}}

\usepackage[dvipsnames, table]{xcolor}
\usepackage{comment}

\usepackage{url}
\usepackage{hyperref}
\hypersetup{
    colorlinks,
    linkcolor={MidnightBlue!75!black},
    citecolor={blue!50!black},
    urlcolor={blue!80!black}
}
\usepackage{graphicx}
\usepackage{booktabs}
\usepackage[counterclockwise]{rotating}
\usepackage{float} 
\usepackage[nolist]{acronym}
\usepackage[numbers, sort]{natbib}

\usepackage{array}
\usepackage{csquotes}

\usepackage{pifont}
\usepackage[T1]{fontenc}
\newcommand{\cmark}{\ding{51}}%
\newcommand{\xmark}{-}%

\usepackage{tabularx}

\usepackage{cleveref}
\crefformat{section}{\S#2#1#3} 
\crefformat{subsection}{\S#2#1#3}
\crefformat{subsubsection}{\S#2#1#3}
\crefrangeformat{section}{\S#3#1#4 to~#5#2#6}
\crefrangeformat{subsection}{\S#3#1#4 to~#5#2#6}
\crefmultiformat{section}{\S(#2#1#3)}{ and~#2#1#3}{,#2#1#3}{ and~#2#1#3}
\crefmultiformat{subsection}{\S#2#1#3}{ and~#2#1#3}{, #2#1#3}{ and~#2#1#3}

\begin{document}

\title{Towards Interpretability in Audio and Visual\\Affective Machine Learning: A Review
\thanks{We gratefully acknowledge funding by the Deutsche Forschungsgemeinschaft (DFG, German Research Foundation): TRR 318/1 2021 – 438445824.}
}
%




\author{\IEEEauthorblockN{David S. Johnson, Olya Hakobyan, and Hanna Drimalla}
\IEEEauthorblockA{\textit{Center for Cognitive Interaction Technology (CITEC)}, \textit{Bielefeld University}\\Bielefeld, Germany}
\IEEEauthorblockA{\{djohnson, ohakobyan, drimalla\}@techfak.uni-bielefeld.de}
}
\maketitle

\begin{abstract}
Machine learning is frequently used in affective computing, but presents challenges due the opacity of state-of-the-art machine learning methods.
Because of the impact affective machine learning systems may have on an individual's life, it is important that models be made transparent to detect and mitigate biased decision making. 
In this regard, affective machine learning could benefit from the recent advancements in \ac{XAI} research. 
We perform a structured literature review to examine the use of interpretability in the context of affective machine learning. We focus on studies using audio, visual, or audiovisual data for model training and identified 29 research articles. Our findings show an emergence of the use of interpretability methods in the last five years. However, their use is currently limited regarding the range of methods used, the depth of evaluations, and the consideration of use-cases. We outline the main gaps in the research and provide recommendations for researchers that aim to implement interpretable methods for affective machine learning.
 
\end{abstract}


\acresetall

\section{Introduction}\label{sec:intro}
Humans communicate their affective state through a rich variety of nonverbal signals, such as facial expressions and vocal tone. These audiovisual signals can be analyzed using machine learning (ML) approaches for tasks like automated affect recognition~\cite{Wang2022_ac_survey, Poria2017_ac-review, Han2021_dl-mentalhealth}.  
Such endeavors have implications for many areas of human life, offering the potential for improved life quality while also bearing risks. 
For instance, automatic recognition of affect and emotions from audiovisual data can enhance human-machine interactions~\cite{Schuller2021_sp-ac} or help develop systems for the diagnosis and management of mental health~\cite{Han2021_dl-mentalhealth}. 
It has already been shown, however, that ML techniques often amplify existing societal biases \cite{Buolamwini2018_} potentially leading to discriminatory interactions or incorrect mental health diagnoses. 
Given the need to provide fair benefits to all users and avoid potential unfair disadvantages, it is critical that decisions made by affective ML systems are transparent and trustworthy to both users of the systems and those affected by their decisions. 

Ensuring transparency in affective ML poses a challenge, especially with the rise of more complex and opaque methods in the field~\cite{Poria2017_ac-review}. For instance, deep-learning methods are becoming more popular due to their high-performance capabilities, yet they are very difficult to interpret. Research from the field of \ac{XAI} aims to address this problem by facilitating the interpretation of models~\cite{Adadi2018_xai_survey}, either by developing inherently interpretable models or through post-hoc explanations~\cite{BarredoArrieta2020_xai_survey}.  Despite their potential, the extent to which interpretability methods are applied in affective machine learning is not clear. 
Even with an awareness of interpretability, adapting these techniques to affective computing may not be straightforward due to different interpretability needs. This calls for a close evaluation of interpretability in affective ML, similar to the review of \Citet{Tjoa2021_medical-xai} on \ac{XAI} in the medical domain.


To our best knowledge, there are no reviews regarding interpretable and explainable affective computing. 
We aim to address this gap by contributing a comprehensive review of interpretability within affective ML for audio and/or visual data, analyzing the techniques employed to obtain interpretable models.
Through this process, we identify the gaps and challenges in the research, and suggest potential research directions to establish interpretable affective ML.

The rest of the paper is organized as follows.  In \cref{sec:interp}, we provide a brief introduction to the definitions and taxonomy of interpretability and explainability. We then introduce the methodology for our review in \cref{sec:methods}, before outlining the results in \cref{sec:results}. The findings are then discussed and summarized in \cref{sec:disc} and \cref{sec:conc}, respectively. Finally, we discuss the ethical impact of our work in \cref{sec:ethics}.

\section{Interpretable AI}\label{sec:interp}
In this section, we provide a brief, non-exhaustive, introduction to the field of interpretable and explainable AI to establish the contextual basis for the relevant concepts. We refer interested readers to the survey of ~\citet{BarredoArrieta2020_xai_survey} for a more complete overview.

Interpretable AI is an emerging and multidisciplinary field without universally accepted definitions of the terms interpretability and explainability, which are often used interchangeably. We follow the taxonomy of \citet{Graziani2022_xai_taxonomy}, however, who consider 
interpretability as a broader term encompassing explainability, and define interpretable AI as a system that is able to communicate its inner workings and decisions in an understandable manner. 
Interpretability may then be achieved by either methods that are intepretable-by-design or that generate explanations post-hoc. 
To this end, we consider interpretability as broader goal and explainability as one potential method to reach this goal.

AI systems that are \textit{interpretable-by-design} include transparent models that can be understood by humans. For example, low-dimensional, or sparse, linear models may be considered transparent since their low-dimensionality affords human-understandable feature weights. Additionally, simple decision-trees that can be parsed by a human allowing them to follow the model's logic would also be considered transparent~\cite{BarredoArrieta2020_xai_survey}.  
Deep learning models with attention mechanisms are also considered by some to have built-in interpretability via their attention weights, although their effectiveness for interpretability is still a topic of debate~\cite{deSantanaCorreia2022_attention, Niu2021_attention}. Attention-based models are considered here to be interpretable-by-design since the attention weights are calculated directly by the model without the need for additional methods after inference to generate an interpretable representation.

\textit{Post-hoc explainability}, on the other hand, describes methods to generate explanations of decisions for models that are not already interpretable.  
Post-hoc methods examine the input data and corresponding predictions using varying techniques to extract what the model has learned and provide users with useful and understandable information about the model~\cite{Lipton2018_mythos-interp}.  One commonly used post-hoc method is feature attribution which generates scores for each feature, indicating their importance toward the final prediction. 

Both interpretable-by-design and post-hoc methods can provide either \textit{global} or \textit{local} explanations~\cite{BarredoArrieta2020_xai_survey, Graziani2022_xai_taxonomy}. Global interpretability approaches target an overall understanding of the model as a whole. For example,  weights from simple linear models can indicate which features (e.g. facial expressions) have the largest influence on an emotion recognition system in general.
By contrast, local interpretability approaches, such as feature attribution, aim to explain the output for a single instance, such as the role of different facial regions towards recognizing the emotion of a particular individual. Local explanations can often be aggregated over a group of instances to provide a global understanding of the model. 

The choice of \textit{input representation} also impacts interpretability~\cite{BarredoArrieta2020_xai_survey}, since input representations differ in their understandability.
For example, high-dimensional vectors of extracted features are often not interpretable since they might not have human-understandable meaning or require expert knowledge to understand. Features such as \acp{FAU}, however, offer a more interpretable representation because they correspond to facial movements understandable by the general population. 
\begin{table*}[t!]
    \centering
    \caption{Interpretability-by-Design in Affective Computing\\(A=Audio, I=Image, V=Video, L=Language)}
    \renewcommand{\arraystretch}{1.35}
    \begin{tabularx}{1.0\textwidth}{
        @{}>{\raggedright\arraybackslash}X
        @{}>{\raggedright\arraybackslash}X
        p{0.025\textwidth}@{}
        p{0.025\textwidth}@{}
        p{0.025\textwidth}@{}
        p{0.03\textwidth}@{}
        >{\raggedright\arraybackslash}X
        >{\raggedright\arraybackslash}X
        p{0.07\textwidth}@{}
        p{0.425\textwidth}@{}
    } 
    \toprule
     \textbf{Article} & \textbf{Task} & \multicolumn{5}{@{}c}{\textbf{Input Modality}}  &  \multicolumn{3}{@{}c}{\textbf{Interpretability}} \\
     \arrayrulecolor{black!40}
     \cmidrule[0.5\cmidrulewidth](lr){3-7}
     \cmidrule[0.5\cmidrulewidth](l){8-10}
     \arrayrulecolor{black}
      &  & \textbf{A} & \textbf{I} & \textbf{V} & \textbf{L} & \textbf{Rep.} & \textbf{Method} & \textbf{Scope} & \textbf{Description}  \\\midrule
    \citet{Seuss2021_} (\citeyear{Seuss2021_}) & Emotional Appraisal & \xmark  & \xmark & \cmark & \xmark  & FAUs & Transparent Model & Global & Feature weights of linear models were used for model analysis and identification of important FAUs. The identified FAUs were compared to the judgments of human experts. \\
    \citet{Cardaioli2022_} (\citeyear{Cardaioli2022_}) & Genuine Expression Detection  &  \xmark & \xmark & \cmark &  \xmark & FAUs & Transparent Model & Global & The rules and weights from decision trees and linear SVMs were used for feature analysis to identify important FAUs indicating genuineness. The rules were visualized as flow charts and models weights visualized using bar charts. \\
    \citet{Jiao2019_} (\citeyear{Jiao2019_}) & Facial Expression Recognition & \xmark & \cmark &  \xmark & \xmark & Raw Images & Attention Mechanism & Local & Attention maps from the attention-based network were used to qualitatively evaluate model interpretability on of a small set of local explanations. \\ 
    \citet{Zhou2020_a} (\citeyear{Zhou2020_a}) & Facial Expression Recognition & \xmark & \xmark & \cmark & \xmark & Facial Landmarks & Graph Network & Local & Saliency maps of the interpreteable graph representation were visualized and used to analyze the model through a qualitative evaluation of a small set of local explanations.\\
    \citet{Anand2021_} (\citeyear{Anand2021_}) & Emotion Recognition & \cmark & \xmark & \xmark & \xmark & Raw Audio & Differential Filter Bank & Global, Local & Proposed a novel deep architecture using differential filterbanks. The cutoff frequencies of each filter were then learned during training, and the frequency responses of the most important filters were visualized to identify important formants for specific emotions.\\
    \citet{Yang2019_} (\citeyear{Yang2019_}) & Facial Expression Recognition & \xmark & \xmark & \cmark & \xmark & Animated GIFs & Attention Mechanism & Local & Facial landmarks were used to generate attention maps during training.  The attention maps of the facial landmarks were used to evaluate the model via a qualitative evaluation on a small set of local explanations.\\
    \citet{Pandit2020_} (\citeyear{Pandit2020_}) & Affect and Pain Recognition & \xmark &  & \cmark & \xmark & FAUs & Transparent Model & Local & Feature weights from extremely shallow CNNs were visualized for interpretability. The weights were used to analyze and improve the model and to perform a qualitative analysis of important features. \\
    \citet{Wu2022_} (\citeyear{Wu2022_}) & Sentiment Analysis & \cmark & \xmark & \cmark & \cmark &  OpenSmile, Embeddings, Word2Vec, & Modality Routing & Global, Local & The capsule network used modality routing to compute the hierarchical importance of each modality and combination of modalities. Interpretability was evaluated with a qualitative review of a few local explanations and with comparison to an attention-based model.\\
    \citet{Wu2021_} (\citeyear{Wu2021_}) & Sarcasm Detection & \cmark & \xmark & \cmark & \cmark & OpenSmile, ResNet, BERT & Attention Mechanism & Local & The attention-based network calculated incongruities between modalities. Attention maps and incongruity scores were visualized and interpretability was analyzed with qualitative review of a small set of local explanations. \\ 
    \citet{Gu2018_} (\citeyear{Gu2018_}) & Sentiment Analysis \& Emotion Recognition & \cmark & \xmark  & \xmark & \cmark & Mel-specs, Word2Vec  & Attention Mechanism & Local & The hierarchical attention-based network used word-level text, audio, and fusion attention modules to identify modality importance for each word. Attention maps were visualized to analyze the model with a qualitative evaluation of a small set of local explanations. \\ 
    \citet{Hemamou2021_} (\citeyear{Hemamou2021_})& Interview Performance & \cmark & \xmark & \cmark & \cmark & eGEMAPS, OpenFace, BERT & Attention Mechanism & Global, Local &  A statistical analysis of attention weights (temporal and modality-level) was performed to validate the attention mechanisms and their interpretability.  \\ %
    \citet{Morales-Vargas2019_} (\citeyear{Morales-Vargas2019_}) & Facial Expression Recognition & \xmark & \xmark & \cmark & \xmark & FAUs & Fuzzy Network & Global & The fuzzy network learned simple FAU-based rule-sets indicating which FAUs contribute to a given model decision. The rule-sets were converted to text for a human understandable explanation. No evaluation of interpretability was performed. \\
    \citet{Browatzki2019_} (\citeyear{Browatzki2019_}) & Facial Expression Recognition & \xmark & \cmark & \xmark & \xmark & Raw Images & Generative Model & Local & Explanations for model decisions were generated by reconstructing input samples from their low-dimensional embeddings.  The interpretability of the generated images was qualitatively evaluated on a small set of local explanations. \\
    \citet{Zhi2020_} (\citeyear{Zhi2020_}) & Facial Expression \& Pain Recognition & \xmark & \xmark & \cmark & \xmark & Raw Video & Sparse Coding & Local & The differentiable programming algorithm learned sparse representations of the face which were visualized to improve the interpretability of the model. The visualizations were used to qualitatively evaluate the model using a small set of local explanation.\\
     \bottomrule
    \end{tabularx}
    \label{tab:inherent-papers}
    \renewcommand{\arraystretch}{1}
\end{table*}

\begin{table*}[t!]
    \centering
    \caption{Post-Hoc Interpretability in Affective Computing\\(A=Audio, I=Image, V=Video, L=Language)}
    \renewcommand{\arraystretch}{1.35}
    \begin{tabularx}{1.0\textwidth}{
        @{}>{\raggedright\arraybackslash}X
        @{}>{\raggedright\arraybackslash}X
        p{0.025\textwidth}@{}
        p{0.025\textwidth}@{}
        p{0.025\textwidth}@{}
        p{0.03\textwidth}@{}
        >{\raggedright\arraybackslash}X
        >{\raggedright\arraybackslash}X
        p{0.07\textwidth}@{}
        p{0.425\textwidth}@{}
    }
    \toprule
     \textbf{Article} & \textbf{Task} & \multicolumn{5}{@{}c}{\textbf{Input Modality}}  &  \multicolumn{3}{@{}c}{\textbf{Interpretability}} \\
     \arrayrulecolor{black!40}
     \cmidrule[0.5\cmidrulewidth](lr){3-7}
     \cmidrule[0.5\cmidrulewidth](l){8-10}
     \arrayrulecolor{black}
      &  & \textbf{A} & \textbf{I} & \textbf{V} & \textbf{L} & \textbf{Rep.} & \textbf{Method} & \textbf{Scope} & \textbf{Description}  \\\midrule
     \citet{Carlini2021_} (\citeyear{Carlini2021_}) & Pain Recognition & \xmark & \cmark & \xmark & \xmark & Raw Images & Feature Attribution & Local & \textit{Integrated Gradients} was used to generate explanations for the decisions of a CNN for assessing neonatal pain in facial images. A few local explanations were qualitatively reviewed to determine important regions of the face for detecting pain.\\
     \citet{Deramgozin2021_} (\citeyear{Deramgozin2021_}) & Facial Expression Recognition & \xmark & \cmark & \xmark & \xmark & Raw Images & Feature Attribution & Global, Local & \textit{LIME} explanations were generated alongside heatmaps of \ac{FAU} intensities present in the image.  The explanation method was qualitatively reviewed using a single sample instance.\\
     \citet{Ghandeharioun2019_} & Facial Expression Recognition & \xmark & \cmark & \xmark & \xmark & Embeddings & Uncertainty & Local & Proposed a model agnostic method to calculate model uncertainty using Monte Carlo dropout. Uncertainty values were shown to be comparable with inter-rater disagreement. The values were used to qualitatively analyze the model with a small set of sample instances. \\
     \citet{terBurg2022_} (\citeyear{terBurg2022_}) & Facial Expression Recognition & \xmark & \cmark & \xmark & \xmark &  Facial Features & Feature Attribution & Local & Geometric facials features were used as an interpretable representation for \textit{SHAP} explanations. The \textit{SHAP} explanations were evaluated by comparing with image-based \textit{GradCAM} explanations in a user study and with a quantitative analysis on explanation fidelity. \\
     \citet{Weitz2019_pain_xai} (\citeyear{Weitz2019_pain_xai}) & Pain Recognition & \xmark & \cmark & \xmark & \xmark & Raw Images & Feature Attribution & Local & \textit{LRP} and \textit{LIME} explanations were compared for explaining the detection of pain from facial images.  Comparison of methods was performed witha qualitative review of a few local explanations. \\
     \citet{Zhou2022_} (\citeyear{Zhou2022_}) & Depression Recognition & \cmark & \xmark & \cmark & \cmark & eGEMAPs, FAUs, Sentiment & Feature Attribution & Global & Feature fusion was used to integrate the modalities before training traditional machine learning methods. The models were analyzed by generating global explanations with \textit{SHAP} to identify important features in the detection of depression and apathy. \\
     \citet{Jain2022_} (\citeyear{Jain2022_}) & Sentiment Analysis & \cmark & \xmark & \cmark & \cmark & COVAREP, OpenFace, GLOVE & Feature Attribution & Global, Local & Modality permutations were used to calculate modality dominance in a co-learning model for multimodal fusion.  \textit{LIME} and \textit{SHAP} were also used to generate local explanations for each modality.  The methods were evaluated with a qualitative review. \\
     \citet{Lorente2021_} (\citeyear{Lorente2021_}) & Facial Expression \& Attention Recognition & \xmark & \cmark & \xmark & \xmark & Raw Images & Feature Attribution & Local & Image-region attributions were generated to explain decisions of a CNN. Interpretability was evaluated by a qualitative review of a few local explanations and the models were analyzed using the explanations to identify important facial regions.\\
     \citet{Asokan2022_} (\citeyear{Asokan2022_}) & Emotion Recognition & \cmark & \xmark & \cmark & \cmark & OpenSmile, Embeddings, Word2Vec & Concepts & Global & \textit{TCAV} was used to compute concept attribution scores for identifying which emotion recognition concepts were most important to the model for each emotion category. The concept models were quantitatively evaluated using concept prediction accuracy. \\
     \citet{Gund2021_} (\citeyear{Gund2021_}) & Facial Expression Recognition & \xmark & \cmark & \xmark & \xmark & Raw Images \& Facial Landmarks & Feature Attribution & Local & Compared the interpretability of class activation maps from a video-based stacked CNN to a temporal convolutional network with facial landmarks sequences as input. The methods were compared by a qualitative review of a few local explanations from each model.\\
     \citet{Rathod2022_} \citeyear{Rathod2022_}) & Facial Expression Recognition & \xmark & \cmark & \xmark & \xmark & Raw Images \& 3D Facial Landmarks & Feature Attribution & Local & Performed a qualitative review of \textit{GradCAM}, \textit{GradCAM++}, and \textit{SoftGrad} explanations across six neural networks and two input representations. The evaluation reviewed a few local explanations for each of the models.\\
     \citet{Boulanger2021_} (\citeyear{Boulanger2021_}) & Affective Engagement Recognition & \xmark & \cmark & \xmark & \xmark & Raw Images & Feature Attribution & Local & Compared local \textit{SHAP} explanations of decisions from a traditional CNN with those of SqueezeNet, a small model optimized for edge devices. The explanations were compared with a qualitative review of a few local explanations from each model. \\
     \citet{Wang2022_} ( \citeyear{Wang2022_}) & Sentiment Analysis & \cmark & \xmark & \cmark & \cmark &  COVAREP, FAUs, GLOVE & Feature Attribution, Modality Dynamics & Global, Local & Proposed an interactive visual analytics system, \textit{M\textsuperscript{2}Lens}, for multimodal explanations of modality dynamics and feature attribution with \textit{SHAP}. \textit{M\textsuperscript{2}Lens} was aimed to help developers understand and debug models, and was evaluated by case studies of three experts.  \\
     \citet{Heimerl2022_a} (\citeyear{Heimerl2022_a}) & Facial Expression Recognition & \xmark & \cmark & \xmark & \xmark & Raw Images & Feature Attribution & Local & A cooperative machine learning annotation tool, NOVA, was proposed for non-expert users that incorporates \textit{LIME} explanations and confidence values to explain model decisions.  NOVA was evaluated by non-expert users for multiple dimensions of interpretability. \\
     \citet{Zhou2020_interp-depress} (\citeyear{Zhou2020_interp-depress}) & Depression Recognition & \xmark & \cmark & \xmark & \xmark & Raw Images & Feature Attribution & Local & Implemented a CNN with global pooling to generate class activation maps to explain the model decisions.  The method was evaluated by qualitative review of a small set of local explanations. \\
     \bottomrule
    \end{tabularx}
    \label{tab:posthoc-papers}
    \renewcommand{\arraystretch}{1}
\end{table*}

\section{Methods}\label{sec:methods}

To identify research on interpretable AI in audio and visual tasks of affective ML, we employed a systematic search via the IEEE, PubMed, and Web of Science electronic databases. We searched each database without explicit start and end dates, on Nov. 11, 2022, for key terms in the title, abstract, and keywords using the query: 
\begin{displayquote}
    {\fontfamily{pcr}\selectfont ("affective computing" OR (("facial expression" OR emotion*) AND (recognition OR detection OR classification OR prediction)) OR ("sentiment analysis" AND multimodal))  AND (xai OR ((explainab* OR interpretab*) AND ("artificial intelligence" OR "AI" OR model* OR "machine learning"))) NOT music}.
\end{displayquote}
The \enquote{*} indicates wildcard search, while the quotations are used for exact search. The query used common terms for machine learning and affective computing, but may have missed some relevant work with different terms or that was not indexed by the included databases.

After removing duplicates, the search resulted in $N=273$ articles. Two of the authors  reviewed and discussed the titles and abstracts of the articles to evaluate the quality of the search results and to finalize the selection criteria. The first inclusion criterion was that the studies employed machine learning methods for the recognition of affect, emotion, facial expression, or sentiment. Studies analyzing affective features (e.g. facial expressions) to predict other variables, such as psychological condition, were included as well. The second inclusion criterion was the use of audio or visual data for model training in an unimodal or multimodal setting. This means that sentiment analysis studies relying solely on textual data were not considered unless they employed multimodal models incorporating audiovisual data. 
Finally, the studies had to list interpretability as one contribution of their research. The exclusion of certain articles was necessary to maintain the focus of the paper, hence studies focusing on music emotion recognition or using exclusively physiological, EEG  or text data were not included.


After this consensus, $N=65$ papers were included for further full-text review. For this iteration, two of the authors (A1 and A2) were randomly assigned half the papers for independent review.  To ensure common ground for inclusion and exclusion in the round, the first $10$ papers were individually reviewed by both authors and subsequently discussed. Papers that merely mentioned interpretability in their abstracts but only briefly touched on the topic in the main text were excluded during this round. In addition, the two authors eliminated papers which, despite including audiovisual data, did not consider behavioral signals (e.g. in a visual emotion analysis classifying the perceived emotional content of an image), focused on non-audiovisual features to explain model predictions (e.g. first impression prediction based on personality traits), or  included intrusive sensors (e.g. motion-capture sensors). Finally, from the various papers that describe the NOVA framework \cite{Heimerl2022_a}, we chose to include the one published in a journal. This resulted in the exclusion of $36$ further papers, for a total of $N=29$ articles that employed interpretable methods for affective ML in the audio and visual domains. All authors then reviewed the final selection of papers to extract information regarding their contributions towards interpretable affective computing.  

For each article we extracted the following information: the main affective ML task, the input modalities and representations, details regarding interpretability (method and scope) and a short description of the method and evaluation. The input modality is specified as the input representation accepted by the model rather than the datasets presented in the paper. For example, in some cases the articles used video datasets but only input individual frames (i.e., images), and not image sequences, into the model. Therefore, we consider these as the image modality.
We classify the input representations as either raw input, features, or embeddings. 
\textit{Raw input} represents unprocessed data such as images, videos or audio signals which are input directly to the model for end-to-end learning approaches.
\textit{Feature} representations are vectors of values that have been extracted from the data, usually through expert-based and domain-specific feature engineering.  
\textit{Embeddings} refers to a form of extracted features that are the learned representations from a model pre-trained on a more general domain-related dataset, rather than being hand-crafted by human experts. 
In the case of features and embeddings, we list specific feature sets when they are considered well-known in affective computing research. 
In some cases, however, the diversity and novelty of methods used makes it challenging to list each method.
\section{Results}\label{sec:results}
We identified $29$ articles employing interpretability methods for affective ML. Although we did not include a date filter, the final selection resulted in publications from the last five years ($2018$-$2022$) indicating the emergence of interpretable affective AI. 
Of these works, $14$ implemented \textit{interpretable-by-design} methods and $15$ used \textit{post-hoc interpretability} methods. Details of the articles for each category are presented in \Cref{tab:inherent-papers}, \textit{interpretability-by-design} and \Cref{tab:posthoc-papers}, \textit{post-hoc interpretability}. 

The affective ML tasks in the reviewed papers range from more traditional applications (affect or emotion recognition, sentiment analysis) to the detection of sarcasm, genuine expressions or depression. Nevertheless, the most common task was the recognition of facial expressions ($N=13$). Most of the models were unimodal, focusing on the visual modality ($N=13$ image, $N=7$ video) with only one paper that relied exclusively on audio data (without the spoken content). There were $N=8$ multimodal methods  using the audio, video and language modalities, except in one case in which only speech data (i.e. audio and language) was included. 
For each modality, there is a range of different input representations that were employed. Typically, visual based methods used raw image or video data ($N=13$) while others extracted features such as \acp{FAU} and facial landmarks ($N=8$). The single audio-only model employed raw audio for model input. Multimodal methods, on the other hand, all relied on extracted features and embeddings.

In the domain of \textit{interpretability-by-design}, \Cref{tab:inherent-papers}, a few articles take the approach of transparent models ($N=3)$. Non-linear models such as deep neural networks with attention mechanisms were used more often ($N=5$). In this case, the attention weights for an input sample provide the user a salient representation of the features most activated in the attention mechanism by that sample. Attention mechanisms can also be implemented as a multimodal fusion module to integrate the input modalities, affording insight into which modality is activated the most for a given decision~\cite{Wu2021_, Gu2018_, Hemamou2021_}. The remaining articles ($N=6$) design models for interpretability using a variety of other novel methods. 
Most of these approaches provide local explanations ($N=11$) while some generate global explanations ($N=6$) (three articles include both global and local explanations). 

In the domain of \textit{post-hoc explainability} methods, \Cref{tab:posthoc-papers}, the most common approach ($N=13$) was the implementation of \textit{feature attribution} methods, such as Locally-Interpretable Model Agnostic Explanations (LIME)~\cite{Deramgozin2021_,Weitz2019_pain_xai,Jain2022_,Heimerl2022_a}, Shapley additive values (SHAP)~\cite{terBurg2022_,Zhou2022_,Jain2022_,Boulanger2021_,Wang2022_}, activation maps (CAM, GradCAM)~\cite{Gund2021_,Rathod2022_}, layer-wise relevance propagation (LRP)~\cite{Weitz2019_pain_xai}, and integrated gradients~\cite{Carlini2021_}. 
Of the remaining articles one proposed a method with Testing Concept Activation Vectors (TCAV)~\cite{Kim2018_} and the other a Monte Carlo dropout method for calculating uncertainty~\cite{Ghandeharioun2019_}. 
Similar to \textit{interpretability-by-design}, the \textit{post-hoc} methods mostly focus on local explanations ($N=13$) but a few implement global explanations ($N=5$) (three articles include both global and local explanations).

Regardless of the interpretability approach, there was an over-reliance on using interpretability methods that visualize feature importance. In the area of \textit{interpretability-by-design} this was done generating attentions maps from the weights of the attention mechanism ($N=5$), visualizing the weights of transparent models ($N=2$) or, in one case, visualizing sparse representations of the input.  Of the \textit{post-hoc explainability} methods, by far the most common approach was to visualize feature attribution scores as either bar charts or saliency maps, visualizing the importance of each pixel towards the model decision ($N=13$).

Our analysis of the research shows that it was taken for granted that the methods implemented were interpretable. In many cases, only a qualitative review of the explanation methods was provided based on a few example explanation visualizations, which were then briefly discussed with a subjective analysis. We identified only $N=4$ works that included users in the evaluation of interpretability, and only one paper that evaluated the validity of the interpretability approach in a quantitative analysis without users.
\section{Discussion}\label{sec:disc}

We reviewed the use of interpretability methods in affective ML, focusing on tasks employing audio and/or visual data. Our results indicate the emergence of both \textit{interpretable-by-design} and \textit{post-hoc explainability} methods over the last years, implemented for a variety of machine learning models with different input types. While these developments are encouraging, the current implementation of these methods is rather limited. We next discuss these limitations and potential solutions for overcoming them. 


In the majority of research reviewed here, interpretability was limited to some variation of saliency maps showing which input features or modalities contributed the most to the decisions of the given model. In addition, we found that very few studies delved deeper into these methods beyond an anecdotal or qualitative review of just a small set of example explanations. This is problematic because the effectiveness of such visualizations for interpretability cannot be assessed without an explicit evaluation.  In fact, it has been shown that relying exclusively on visualizations can be misleading and that many saliency methods perform similar to edge detectors that are model- and data-independent~\cite{Adebayo2018_}. Likewise, the reliability of attention weights for deriving model decisions has been questioned~\cite{Niu2021_attention}. \textit{Hence, we recommend performing systematic evaluation and comparison of different interpretability methods to ensure that the explanations are understandable and make sense for the given application.}

It is worth noting that even a carefully chosen interpretability method may only show \textit{what} features are important but not necessarily explain \textit{why} these features have a greater impact on the model's decisions than the others, resulting in an interpretability gap that must be filled by the end-user~\cite{Ghassemi2021_}. 
The severity of the interpretability gap can vary based on the type of input representation. For example, studies that used raw input for image data can generate salience maps that are more easily interpretable, compared to other input types, such as audio spectrograms which require expert knowledge to achieve a level of interpretability. 
Similarly, extracted features, and especially embeddings, are often not intuitive, calling for a careful consideration of input representations when developing interpretable affective computing systems. \textit{It may be advisable to consider interpretable features whenever possible or alternative methods to feature attribution.} For instance, concept activation vectors (CAVs)~\cite{Kim2018_, Asokan2022_} provide explanations in terms of task specific high-level concepts that have been designed by the developer to be understood by the model's user group, instead of relying on low-level feature vectors. A more detailed description of alternative post-hoc explainability methods is available in the review of \citet{BarredoArrieta2020_xai_survey}.  

One aspect that may complicate relating input representations to model decisions is the use of multimodal approaches. Unlike in unimodal systems, understanding model decisions in these cases requires the consideration of a wider array of inputs from multiple modalities. Many of the reviewed methods focused on identifying dominant modalities, but largely neglected the interactions between the modalities. \textit{Therefore, we recommend analyzing the interactions between modalities similar to the methods proposed by \citet{Wang2022_} for M\textsuperscript{2}Lens, indicating how different modalities can complement, dominate or conflict with each other}.

 Another challenge for multimodal scenarios is that availability of established interpretability methods differ between modalities. The methods used in the visual domain are more advanced in this regard compared to audio, where the interpretability research is still in its early stages. \textit{Therefore, more research into interpretability for audio modalities might be a promising direction, including the development of new interpretable features similar to \acp{FAU} for facial expressions}. 

In terms of scope, local explanations outnumbered global ones. While global explanations can be useful for identifying stereotypical features of affective expression, they may not capture finer variability between people and different situations. In particular, inter-personal level emotion expression can vary based on characteristics, such as gender or culture~\cite{Hall2019_}. At an intra-personal level, people's behavior might change depending on the situation, e.g. whether or not they are stressed. \textit{In this way, global explanations may be used to understand and identify biases and generalizations of models towards different groups, while local explanations may be used to provide an understanding of how intra- and inter-personal variations affect model decisions for specific individuals.}

Selecting an effective interpretability method for an explanation context requires an understanding of whom the explanation is for, but \ac{XAI} researchers often fall short of considering the potential end-users~\cite{Miller2017_}. 
For this reason, it is recommended that XAI methods should be developed with the goal of being understandable to end-users and hence be evaluated with human behavioral studies.
However, we found it to be rare that researchers even mentioned the intended recipients of the explanations. Furthermore, only four of the reviewed papers performed user-based evaluations~\cite{Seuss2021_, terBurg2022_, Wang2022_, Heimerl2022_a}.
In line with recent calls for more rigorous~\cite{Doshi-Velez2017_iml} and human-based~\cite{Miller2017_} evaluation of interpretability, \textit{we suggest that affective computing researchers take a more human-centered approach and consider the contexts in which explanations will be situated, including both the application tasks and users.} For example, explanations during an interaction between an affective social robot and a user with autism will have different needs than explanations geared towards an interviewer aiming to understand the behavioral analysis of an interview candidate.
 

Following a human-centered approach~\cite{Miller2017_}, explanations can be seen as conversations and machine explanations should follow their maxims.  This does not mean machine explanations have to be textual, but rather that they should have the properties found in human conversation.
\citet{Rohlfing2020_} build on this and suggest that explanations are a social practice and should be co-constructed between system and user. 
This requires supporting a variety of explanation methods and knowledge of how they apply to different situations.  In our analysis, only the work of \citet{Wang2022_} proposed interactive explanations, but this was limited to specific \ac{XAI} methods and did not allow the system to adapt the explanation methods to the users needs. 
To this end, \textit{we suggest researchers should focus on implementing and evaluating a breadth of interpretability methods, beyond just feature importance, in different affective computing contexts}. For example, researchers may want to consider counterfactual explanations as they are more congruent with human reasoning by considering how a model's decision change under different conditions~\cite{Miller2017_}.
\section{Conclusion}\label{sec:conc}
We welcome the use of interpretability methods within the field of affective machine learning. However, interpretable affective ML is still a young field with a limited breadth and evaluation of methods, calling for increased clarity regarding the context of the explanations and the effectiveness of different interpretability methods within stated contexts. Our analysis shows that  affective ML could benefit from explicit consideration and discussion of the interpretability approach, such as the reliability of the used method, the interpretability of input representations and the choice of interpretability scope. Finally, the real end-users of affective ML systems should be considered when designing interpretable systems. While our work outlined the main developments and limitations, future work is needed for a more in-depth analysis of the use of interpretability methods for different contexts of different affective ML tasks. 



\section{Ethical Impact Statement}\label{sec:ethics}
With our review and recommendations on the use of interpretability methods in affective ML, we hope to contribute to the development of more transparent and ethical systems. Nevertheless, it is important to approach our recommendations with caution. Specifically, we explicitly emphasize that implementing interpretable methods, even if approached in a rigorous manner, does not guarantee ethical systems. For example, providing explanations has been shown to increase automation bias leading to users over-trusting model decisions~\cite{Kaur2020_interp_interp} potentially leading to adversarial implementations of interpretability. It is necessary then to consider the entire pipeline of the machine learning approach in an interdisciplinary setting, considering all steps from data collection to deployment.

\begin{acronym}[XAI-AC]
    \acro{FAU}{facial action unit}
    \acro{HCI}{human-computer interaction}
    \acro{XAI}{explainable artificial intelligence}
\end{acronym}

\bibliographystyle{IEEEtranN}
\bibliography{IEEEabrv,xai_ac_review}

\end{document}